\documentclass[runningheads]{llncs}

\usepackage{hyperref}
\usepackage{graphicx}
\usepackage{amsmath}
\usepackage{amssymb}
\usepackage{xspace}
\usepackage{mathptmx}
\usepackage{stmaryrd}

\newcommand\cupEq{\protect{~\cup{\kern -0.5em}=~}}
\newcommand\ncupEq{\protect{~\#{\kern -0.3em}\cup{\kern -0.5em}=~}}
\newcommand{\ELp}{$\mathcal{E \kern -0.2em L}^+$} 
\newcommand{\ELpp}{$\mathcal{E \kern -0.2em L}^{++}$}

\newcommand{\Protege}{Prot\'{e}g\'{e}\xspace}

\title{Modeling OWL with Rules: The ROWL \Protege Plugin}
\titlerunning{Modeling OWL with Rules: The ROWL Prot\'eg\'e Plugin}

\author{Md. Kamruzzaman Sarker\inst{1} \and David Carral\inst{1}  \and \\ Adila A. Krisnadhi\inst{1,2} \and Pascal Hitzler\inst{1}}
\institute{Wright State University, OH, USA \and Universitas Indonesia, Depok, Indonesia}

\authorrunning{Sarker, Carral, Krisnadhi, Hitzler}

\begin{document}

\maketitle

\begin{abstract}
  In our experience, some ontology users find it much easier to convey logical statements using rules rather than OWL (or description logic) axioms.
  Based on recent theoretical developments on transformations between rules and description logics, we develop ROWL, a \Protege plugin that allows users to enter OWL axioms by way of rules; the plugin then automatically converts these rules into OWL DL axioms if possible, and prompts the user in case such a conversion is not possible without weakening the semantics of the rule.
\end{abstract}


\section{Motivation}\label{sec:motivation}

It has long been argued, that rules are much more intuitive and easier to master than description logics, in terms of what their intended meaning is. We find this substantiated throughout our
experiences as teachers and as ontology modelers which frequently work with domain experts.

To give just a simple example: The exact semantics behind a logical axiom such as 
$$\textsf{Journal} \sqsubseteq \forall\textsf{publishedBy}.\textsf{Organization}$$
in our experience often remains somewhat unclear even for people with significant exposure to ontologies and ontology modeling. 
On the other hand, a rule such as
$$\textsf{Journal}(x) \wedge \textsf{publishedBy}(x,y) \to \textsf{Organization}(y)
$$
is rather intuitive for most in its meaning, and can be both produced and processed much more readily. 

The axiom and the rule just given are of course logically equivalent.\footnote{When we interpret the rule in the sense of first-order predicate logic, i.e., according to the open world semantics.} In
fact many OWL axioms can be expressed equivalently as rules, which are, arguably, easier to understand and to produce.

As a consequence of these observations, we have produced a Prot\'eg\'e plugin which accepts rules as input, and adds them as OWL axioms to a given ontology, provided the rule is expressible by an
equivalent set of such axioms. In case the rule is not readily transferable, the user is prompted and asked how to translate the rule, as there are different options how to do it in such cases. More
information about the plugin can be found at \url{http://daselab.org/content/modeling-owl-rules}.



\section{Rules-to-OWL Transformation}\label{sec:transform}

In this section, we provide some examples of translations of rules into OWL axioms  in an attempt to convey an intuitive understanding of our transformation.
For a formal and complete of such procedure we
refer the reader to \cite{DBLP:conf/esws/MartinezH12}. 
Note that, as opposed to \cite{DBLP:conf/esws/MartinezH12}, we do not consider rules in our implementation that would require the use of role
conjunction, as this is a logical constructor not currently allowed in OWL.

The following rule can be used to characterize all individuals taking courses and working for a department as student workers.
\begin{align}
\textsf{attends}(x, y) \wedge \textsf{Course}(y) \wedge \textsf{worksFor}(x, z) \wedge \textsf{Dept}(z) \to \textsf{StudentWorker}(x) \label{firstRule}
\end{align}

We transform this rule into a DL axiom via a series of equivalence preserving transformations.
First, we detect that both $y$ and $z$ are variables that can be ``rolled up,'' as they only occur in a single object property.
Thus, these variables can be sequentially removed from the rule, resulting in the following:
\begin{align*}
\exists \textsf{attends}.\textsf{Course}(x) \wedge \textsf{worksFor}(x, z) \wedge \textsf{Dept}(z) &\to \textsf{StudentWorker}(x) \\
\exists \textsf{attends}.\textsf{Course}(x) \wedge \exists \textsf{worksFor}.\textsf{Dept}(x) &\to \textsf{StudentWorker}(x)
\end{align*}

\noindent Furthermore, we can unify all unary atoms of the form $C(x)$, i.e., sharing the same variable $x$, yielding:
\begin{align*}
(\exists \textsf{attends}.\textsf{Course} \sqcap \exists \textsf{worksFor}.\textsf{Dept})(x) &\to \textsf{StudentWorker}(x)
\end{align*}

\noindent The previous rule can then be directly translated into OWL as the following axiom:
\begin{align*}
  \exists \textsf{attends}.\textsf{Course} \sqcap \exists \textsf{worksFor}.\textsf{Dept} \sqsubseteq \textsf{StudentWorker}
\end{align*}

For the next example, we have the following rule, which specifies that ``all mice are smaller than all elephants.''
\begin{align*}
\textsf{Mouse}(x) \wedge \textsf{Elephant}(y) \to \textsf{smallerThan}(x, y)
\end{align*}

Translating such a rule into OWL requires us to first connect the variables in the body.  We do so by adding atoms of the form $U(t, u)$ with $U$ the universal property, i.e.,
\textsf{owl:topObjectProperty}.
\begin{align*}
\textsf{Mouse}(x) \wedge U(x, y) \wedge \textsf{Elephant}(y) \to \textsf{smallerThan}(x, y)
\end{align*}

The previous rule can be directly translated into OWL resulting in the following three axioms where $R_{\mathsf{Mouse}}$ and $R_{\mathsf{Elephant}}$ are fresh object properties not previously
occurring in the ontology:
\begin{gather*}
  \textsf{Mouse} \sqsubseteq \exists R_\textsf{Mouse}.\textsf{Self} \\
  \textsf{Elephant} \sqsubseteq \exists R_\textsf{Elephant}.\textsf{Self} \\
  R_{\textsf{Mouse}} \circ U \circ R_{\textsf{Elephant}} \sqsubseteq \textsf{smallerThan}
\end{gather*}

Certain rules cannot be expressed in OWL employing our approach.
For example, the following rule, which characterizes the set of individuals taught by their own uncle, cannot be translated by our approach.
\begin{align*}
\textsf{hasFather}(x, y) \wedge \textsf{hasBrother}(y, z) \wedge \textsf{taughtBy}(x, z) \to \textsf{TaughtByUncle}(x)
\end{align*}
Note that, such a rule cannot be reduced in the same way as rule \label{firstRule}, since every variable occurs in at least two atoms with object properties as predicates.

In cases, such as the previous one, in which a rule cannot be translated into OWL using a set of DL axioms, our implementation will suggest several options to translate such rule using nominal schemas
\cite{KrotzschMKH11}.  The chosen option by the user will be recorded in an annotation which will be added to the rule.  As of right now, there is no syntax for nominal schemas in OWL and thus, we
have decided that an annotation is the best way to convey such information.

\section{Plugin Description and Features}\label{sec:system}




\begin{figure}[tp]
  \centering
  \includegraphics[width=\textwidth]{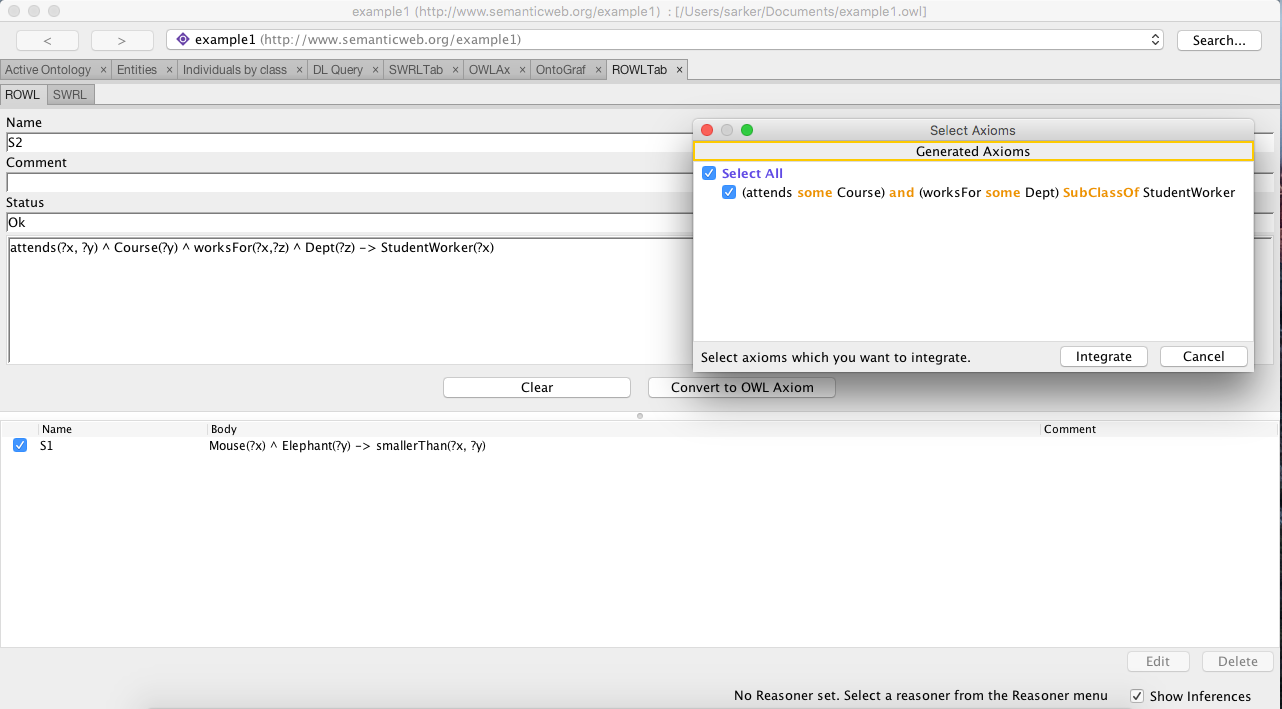}
  \caption{The ROWL interface. The pop-up window appears after clicking ``Convert to OWL Axiom'' button and the transformation is successful.}
  \label{fig:rowl}
\end{figure}

Figure \ref{fig:rowl} depicts the user interface of the ROWL plugin. This plugin is implemented on top of \Protege's SWRLTab plugin
implementation
and thus, it borrows the pretty much SWRLTab user interface for entering rules as
input. As seen in the figure, the plugin consists of two tabs: ROWL and SWRL. The latter is really SWRLTab input interface, while the former is a modification of the latter where we add ``Convert
to OWL Axioms'' button. A user can enter a rule in ROWL tab using the standard SWRL syntax, e.g.: 

\begin{center}
  \textsf{attends}(?x, ?y) \^{} \textsf{Course}(?y) \^{} \textsf{worksFor}(?x, ?z) \^{} \textsf{Dept}(?z) -$>$ \textsf{StudentWorker}(?x)
\end{center}

\noindent When the ``Convert to OWL Axiom'' button is clicked, ROWL will attempt to apply the rules-to-OWL transformation described in the previous section to the given rule. If successful, a pop-up
will appear displaying one or more OWL axioms resulted from the transformation, presented in Manchester syntax. These axioms  can then be integrated into the active ontology. 

If the given rule cannot be transformed into OWL axiom, ROWL will prompt the user if they still want to insert the rule into the ontology as an SWRL rule with annotation. If the user agrees, ROWL will
switch to its SWRL tab and proceed in the same way as adding a rule via the original SWRLTab. Note that ROWL is separate from the original SWRLTab, hence any SWRL rule added via ROWL will not affect
rules added through the original SWRLTab.

Note that, once the axioms generated from rules are added into the ontology, the plugin does not provide a way to undo such modification and recover the original rule from which the axioms were generated.
That is, when a user enters a rule through this plugin and converts it to OWL axioms, the active ontology is either augmented with the generated OWL axioms or SWRL rules.
To implement this feature, we would need a way to record which axioms were generated from which rules.
This will be considered as part of future development of this plugin.

Finally, a feature of ROWL not found in SWRLTab is the possibility to automatically add declarations for classes and properties if the inserted rule contain classes or properties not yet defined in the ontology.
For example, in the rule above, the original SWRLTab requires that \textsf{attends} and \textsf{worksFor} to be already defined as object property, while \textsf{Course}, \textsf{Dept},
and \textsf{StudentWorker} as class in the ontology.
This would add a little bit more freedom for the user to enter any rule (s)he wishes during modeling because (s)he does not need to first exit the
plugin and declare the classes and properties directly in \Protege.





\bigskip

\noindent\emph{Acknowledgements.} This work was supported by the National Science Foundation under award 1017225 \emph{III: Small: TROn -- Tractable Reasoning with Ontologies}.

\bibliographystyle{splncs03}
\bibliography{all}

\end{document}